\documentclass[12pt]{article}
\usepackage{amsmath, amsthm, amssymb}
\usepackage{latexsym}
\usepackage{exscale}
\usepackage{fontenc}
\usepackage{graphicx}
\usepackage{epsfig}

\def\ba{{\mathbf a}}
\def\bi{{\mathbf i}}
\def\bj{{\mathbf j}}
\def\bx{{\mathbf x}}
\def\bf{{\mathbf f}}
\def\bo{{\mathbf o}}
\def\bp{{\mathbf p}}
\def\by{{\mathbf y}}

\def\bA{{\mathbf A}}
\def\bB{{\mathbf B}}

\def\bV{{\mathbf V}}
\def\bL{{\mathbf L}}

\begin{document}
\title{Visual Grouping by Neural Oscillators}
\author{Guoshen Yu~\footnote{CMAP, Ecole Polytechnique, 91128 Palaiseau Cedex, France. (email: yu@cmap.polytechnique.fr)}
\and Jean-Jacques Slotine~\footnote{NSL, Massachusetts Institute of Technology, Cambridge, MA 02139, USA. (email: jjs@mit.edu)}}

\maketitle

\begin{abstract}
Distributed synchronization is known to occur at several scales in the
brain, and has been suggested as playing a key functional role in
perceptual grouping. State-of-the-art visual grouping algorithms,
however, seem to give comparatively little attention to neural
synchronization analogies. Based on the framework of concurrent
synchronization of dynamic systems, simple networks of neural
oscillators coupled with diffusive connections are proposed to solve
visual grouping problems.  Multi-layer algorithms and feedback
mechanisms are also studied. The same algorithm is shown to
achieve promising results on several classical visual grouping
problems, including point clustering, contour integration and image
segmentation.
\end{abstract}

\section{Introduction}

\begin{figure}[htbp]
\begin{center}
\begin{tabular}{ccc}
\includegraphics[width=4.2cm]{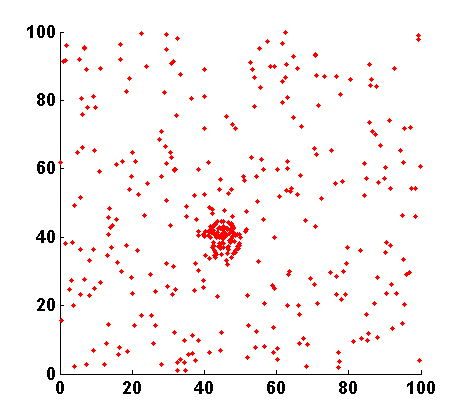}
&
\includegraphics[width=4cm]{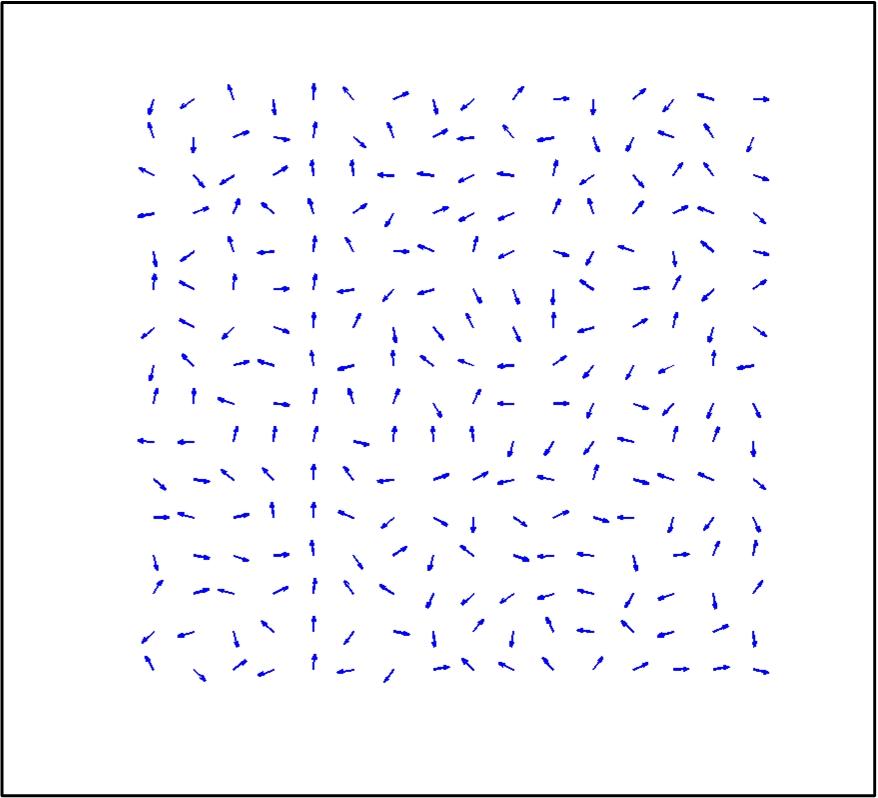}
& \hspace{1ex}
\includegraphics[width=3.8cm]{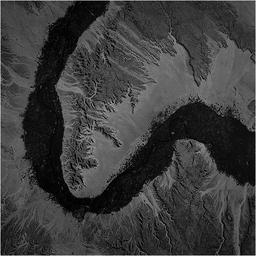}
\\
\end{tabular}
\end{center}
\caption{Left: a cloud of points in which a dense cluster is
embedded. Middle: a random direction grid in which a vertical
contour is embedded. Right: an image in which a river is
embedded.} \label{fig:visual:examples}
\end{figure}

Consider Fig.~\ref{fig:visual:examples}. Why do we perceive in these
visual stimuli a cluster of points, a straight contour and a river?
How is the identification performed between a subgroup of stimuli and
the perceived objects?  These classical questions can be addressed
from a variety of point of views, both biological and mathematical.
This paper develops new grouping algorithms in the
biologically-inspired framework of distributed oscillator
synchronization.

Many physiological studies, e.g.~\cite{Field93, hess2003cia,
lee2003cev}, have shown evidence of grouping in visual cortex.
Gestalt psychology~\cite{wertheimer23untersuchungen, metzger2006ls,
kanizsa1996gvm, desolneux05computational}, an attempt to formalize
the laws of visual perception, addresses some grouping principles
such as proximity, good continuation and color constancy, in order
to describe the construction of larger groups from atomic local
information in the stimuli.

In computer vision, various mathematical frameworks have been
suggested for grouping different visual qualities. Besides some
classical clustering algorithms~\cite{dubes1988acd} such as
k-means~\cite{MacQueen67}, graph-based methods have been proposed
for point clustering~\cite{OlsonSCGP2005, shi2000nca}. Geometrical
grouplets~\cite{mallat08grouplets} consider contour grouping in the
framework of harmonic analysis. The \textit{ a contrario}
school~\cite{cao05:_detecting_groups, desolneux03grouping,
desolneux03computational, desolneux05computational} applies
probabilistic approaches to address the perceptual grouping such as
point clustering and contour detection. Variational
formulations~\cite{mumford1989oap,morel1995vmi,aubert2002mpi},
Markov Random Fields~\cite{geman1987srg} and graph
cuts~\cite{shi2000nca} have been applied to perform image
segmentation. These computer vision approaches have achieved
important success. However, as most of them have been proposed under
a specific motivation (and often without much interest for
biological analogy), applications of each algorithm are usually
limited in grouping based on one specific quality.

In the brain, at a finer level of functional detail, the distributed
\textit{synchronization} known to occur at different scales has been
proposed as a general functional mechanism for perceptual
grouping~\cite{buzsaki2006rb,singer1995vfi}.  In computer vision,
comparatively little attention has been devoted to exploiting
neural-like oscillators in visual grouping. Wang and his colleagues
have performed very innovative work using oscillators for image
segmentation~\cite{wang1997isb, Liu99,chen2002dcn} and have extended
the scheme to auditory segregation~\cite{brown1997mps, wang1999ssi,
wang2008ocm}. They constructed oscillator networks with local
excitatory lateral connections and a global inhibitory connection.
Due to the large complexity, for real images a segmentation algorithm
essentially without oscillators was abstracted from the underlying
oscillatory dynamics~\cite{chen2002dcn}. Li has proposed elaborate
visual cortex models with oscillators~\cite{li1998nmc, li1999vsc,
li2000pas, li2001cda} and applied them on lattice drawings. Kuzmina
and his colleagues~\cite{kuzmina2001ton, kuzmina2004ons} have
constructed a simple self-organized oscillator coupling model, and
applied it on synthetic lattice images as well. Faugeras et al., have
started studying oscillatory neural mass models in the contexts of
natural and machine vision~\cite{Faugeras07}.

In this paper we propose a simple and general neural oscillator
algorithm for visual grouping, based on diffusive
connections~\cite{Pham07}. We use full-state neural oscillator models
rather then phase-based approximations. The key to our approach is to
embed the desired grouping properties in the \textit{couplings}
between oscillators. This allows one to exploit existing results on
visual grouping and Gestalt while at the same taking advantage of the
flexibility and robustness afforded by synchronization mechanisms.
Synchronization of oscillators induces perceptual grouping while
desynchronization leads to segregation. Multi-layer networks with
feedback are introduced. Applications to point clustering, contour
integration, and segmentation of synthetic and real images are
demonstrated.  A recent study of stable concurrent synchronization of
neural oscillators~\cite{Pham07} provides a general analysis tool to
model the associated nonlinear dynamics and study their convergence
properties.

Section~\ref{sec:model} introduces a basic model of neural oscillators
with diffusive coupling connections, studies its stability, and
proposes a general visual grouping algorithm.
Sections~\ref{sec:point:clustering},~\ref{sec:contour:integration}
and~\ref{sec:image:segmentation} describe in detail the neural
oscillator solutions for point clustering, contour integration and
image segmentation and show a number of examples.
Section~\ref{sec:conclusion} presents brief concluding remarks.

\section{Model and Algorithm}
\label{sec:model} The model is a network of neural oscillators
coupled with diffusive connections. Each oscillator is associated to
an atomic element in the stimuli, for example a point, an
orientation or a pixel. Without coupling, the oscillators are
desynchronized and oscillate in random phases. Under diffusive
coupling with the coupling strength appropriately tuned, they may
converge to multiple groups of synchronized elements. The
synchronization of oscillators within each group indicates that
perceptual grouping of the underlying stimulative atoms, while the
desynchronization between groups suggest group segregation.

The next section describes how to construct the neural oscillator
networks and shows the stability and convergence properties of the
system. A general visual grouping algorithm is proposed at the end.

\subsection{Neural Oscillators}
\label{subsec:oscillators}

We use a modified form of FitzHugh-Nagumo neural
oscillators~\cite{FitzHugh61,Nagumo62}, similar
to~\cite{Liu99,chen2002dcn},
\begin{eqnarray}
\label{eqn:oscillator:FN} \dot{v}_i & = & 3 v_i - v_i^3 - v_i^7 + 2
- w_i + I_i \\
\dot{w}_i & = & c [\alpha (1+\tanh(\beta v_i)) - w_i]
\end{eqnarray}
where $v_i$ is the membrane potential of the oscillator, $w_i$ is an
internal state variable representing gate voltage, $I_i$ represents
the external current input, and $\alpha$, $\beta$ and $c$ are
strictly positive constants (we use $\alpha = 12; c = 0.04; \beta =
4$). When the input $I_i$ exceeds a certain threshold value, the
neural oscillator oscillates, the trace of membrane potential $v_i$
being plotted in Fig.~\ref{fig:synch:2osc}-a. Other spiking
oscillator models can be used similarly. In the neural oscillator
networks for visual grouping, each oscillator is associated to an
atomic element in the stimuli.

\subsection{Diffusive connections}\label{subsecdiffusive}

Oscillators are coupled using diffusive connections with
Gaussian-tuned gains to form networks.

Let us denote by $\mathbf{x_i}=[v_i, w_i]^T$ the state vectors of the
oscillators introduced in section~\ref{subsec:oscillators}, each with
dynamics $\dot{\mathbf{x}}_i = \mathbf{f} (\bx_i, t)$. A neural
oscillator network is composed of $N$ oscillators, connected with
diffusive coupling~\cite{wang05a}
\begin{equation}
\label{eqn:dif:connection} \dot\bx_i = \bf (\bx_i, t) + \sum_{i \neq
j} k_{ij} (\bx_j - \bx_i),~~~~i=1,\ldots,N
\end{equation}
where $k_{ij}$ is the coupling strength.

\begin{figure}[htbp]
\begin{center}
\begin{tabular}{cc}
\includegraphics[width=6cm]{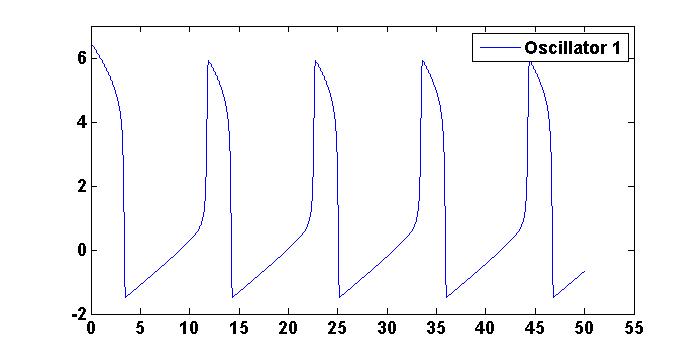}
&
\includegraphics[width=6cm]{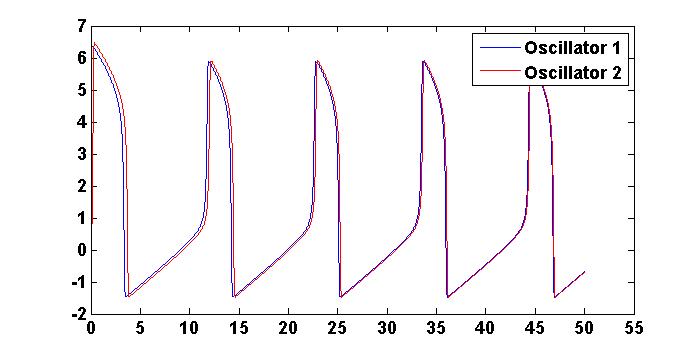}
\\
\textbf{a} & \textbf{b} \\
\end{tabular}
\end{center}
\caption{\textbf{a.} a single oscillator. \textbf{b.}
synchronization of two oscillators coupled through diffusive
connections. The two oscillates start to be fully synchronized at
about $t=30$.} \label{fig:synch:2osc}
\end{figure}

Oscillators $i$ and $j$ are said to be synchronized if $\bx_i$
remains equal to $\bx_j$. Once the elements are synchronized, the
coupling terms disappear, so that each individual elements exhibits
its natural, uncoupled behavior, as illustrated in
Fig.~\ref{fig:synch:2osc}. It is intuitive to see that a larger
$k_{ij}$ value facilitates and reinforces the synchronization
between the oscillators $i$ and $j$  (refer to Appendix for more
details).

The key to apply neural oscillators with diffusive connections to
visual grouping is to tune the coupling so that the oscillators
synchronize if their underlying atoms belong to the same visual
group and desynchronize otherwise. According to Gestalt
psychology~\cite{wertheimer23untersuchungen,kanizsa1996gvm,metzger2006ls},
visual stimulus atoms of similarity or proximity tend to be grouped
perceptually. This suggests that the coupling between the neural
oscillators should be reinforced if they have similar stimuli. Such
coupling can be implemented by the Gaussian tuning
\begin{equation}
\label{eqn:coupling:strength} k_{ij} =
e^{\frac{-|s_i-s_j|^2}{\beta^2}}.
\end{equation}
where $s_i$ and $s_j$ are stimuli of the two oscillators, for
example position for point clustering, orientation for contour
integration and grey-level for image segmentation, and $\beta$ is a
tuning parameter. Due to its good properties such as smoothness,
Gaussian tuning has been applied in various applications such as
image denoising~\cite{Buades05,Buades07},
segmentation~\cite{shi2000nca} and recognition~\cite{serre2007ror}.

\subsection{Generalized diffusive connections}
\label{subsubsec:generalized}

The visual cortex is hierarchical. Higher-level layers have smaller
dimension than lower-level layers, as information going from bottom
to top turns from redundant to sparse and from concrete to
abstract~\cite{rao1999pcv}.

In a feedback hierarchy, generalized diffusive connections correspond
to achieving consensus between multiple processes of different
dimensions.  Implementation of the hierarchy involves connecting two
or more oscillator subnetworks of different sizes.  Oscillator
networks of sizes $N_1$ and $N_2$ can be connected with generalized
diffusive connections~\cite{wang05b,Pham07}:
\begin{eqnarray}
\dot{\mathbf{x}}_1 & = & \mathbf{f}_1 (\bx_1, t) + k_1 \bA^T (\bB
\bx_2
- \bA \bx_1) \label{eqn:generalized:l1} \\
\dot{\mathbf{x}}_2 & = & \mathbf{f}_2 (\bx_2, t) + k_2 \bB^T (\bA
\bx_1 - \bB \bx_2 ), \label{eqn:generalized:l2}
\end{eqnarray}
where $\bf_1$ and $\bf_2$ are dynamics of two networks of sizes
$N_1$ and $N_2$ whose state vectors are respectively $\bx_1$ and
$\bx_2$, $\bA$ and $\bB$ are coupling matrices of appropriate sizes
$M \times N_1$ and $M \times N_2$, $k_1$ and $k_2$ are the coupling
strengths.

For appropriate choices of dynamics, once the two layers are
synchronized, i.e.~$\bA \bx_1 = \bB \bx_2$, the coupling terms
disappear, so that each layer exhibits its natural behavior as an
independent network, as in the case of diffusive connections.

\begin{figure}[htbp]
\begin{center}
\begin{tabular}{c}
\includegraphics[width=10cm]{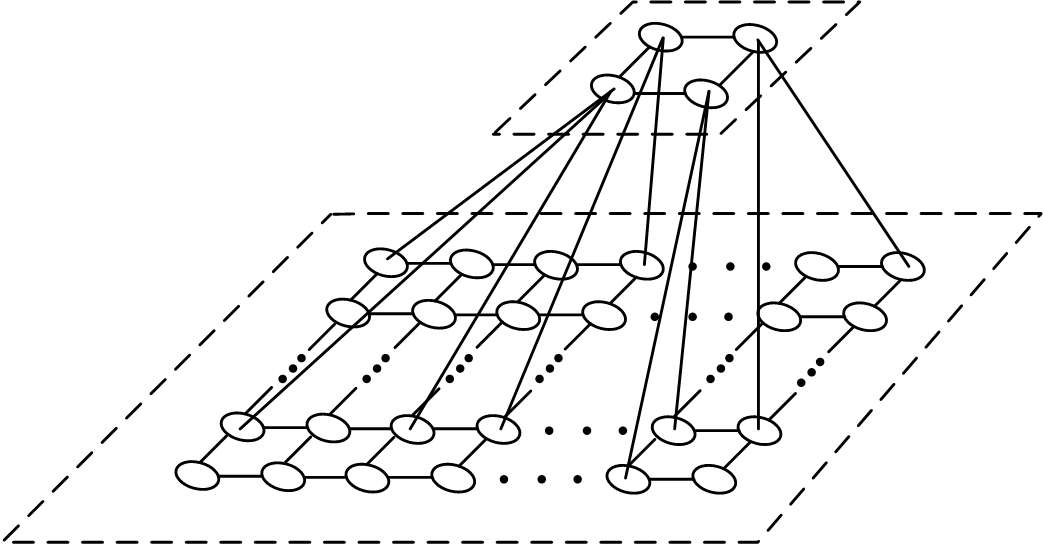}
\\
\end{tabular}
\end{center}
\caption{Two networks (top and bottom) of different dimensions are
connected with \textit{generalized diffusive connections}. }
\label{fig:extension:diffusive}
\end{figure}

\subsection{Concurrent Synchronization and Stability}

In perception, fully synchronized elements in each group are bound,
while different groups are segregated. Concurrent synchronization
analysis provides a mathematical tool to study stability and
exponential convergence properties in this context.

In an ensemble of dynamical elements, concurrent synchronization is
defined as a regime where the whole system is divided into multiple
groups of fully synchronized elements, but elements from different
groups are not necessarily synchronized~\cite{Pham07}.  Networks of
oscillators coupled by diffusive connections
(section~\ref{subsecdiffusive}) or generalized diffusive connections
(section~\ref{subsubsec:generalized}) are specific cases of this
general framework.

Recall that a subset of the global state space is called \textit{
invariant} if trajectories that start in that subset remain in that
subset. In our synchronization context, the invariant subsets of
interest are linear subspaces, corresponding to some components of
the overall state being equal or verifying some linear relation.
Concurrent synchronization analysis quantifies stability and
convergence to invariant linear subspaces.  Furthermore, a property
of concurrent synchronization analysis, which turns out to be
particularly convenient in the context of grouping, is that the
actual invariant subset itself need not be know \textit{a priori}
to guarantee stable convergence to it.

Finally, concurrent synchronization may first be studied in an
idealized setting, e.g., with exactly equal inputs to groups of
oscillators, and noise-free conditions. This allows one to compute
minimum coupling gains to guarantee global exponential convergence to
the invariant synchronization subspace.  \textit{Robustness} of
concurrent synchronization, a consequence of its exponential converge
properties, allows the qualitative behavior of the nominal model to be
preserved even in non-ideal conditions. In particular, it can be shown
and quantified that for high convergence rates, actual trajectories
differ little from trajectories based on an idealized model.

A more specific discussion of stability and convergence is given in
the appendix. The reader is referred to~\cite{Pham07} for more details
on the analysis tools.

\subsection{Visual Grouping Algorithm}
\label{subsec:algo} The basic visual grouping algorithm proceeds in
the following steps.
\begin{enumerate}
\item Construct a neural oscillator network. Each oscillator is
associated to one atom in the stimuli. Oscillators are connected with
diffusive connections (\ref{eqn:dif:connection}) or generalized
diffusive connections (\ref{eqn:generalized:l1},
\ref{eqn:generalized:l2}) using the Gaussian-tuned gains
(\ref{eqn:coupling:strength}).
\item The oscillators converge to concurrently synchronized
groups in the so-constructed network.
\item Identify the synchronized oscillators and equivalently the visual
groups. A group of synchronized oscillators indicates that the
underlying visual stimulative atoms are perceptually grouped.
Desynchronization between groups suggest that the underlying
stimulative atoms in the two groups are segregated.
\end{enumerate}

Traces of synchronized oscillators coincide in time, while those of
desynchronized groups are separated~\cite{DLWang05}.  The
identification of synchronization in the oscillation traces (as
illustrated in the example of
Fig.~\ref{fig:clustering:points:2:groups}-b) can be realized by
thresholding the correlation of the traces or by simply applying a
clustering algorithm such as $k$-means.

The following sections detail neural oscillator solutions for
three visual grouping problems, namely point clustering, contour
integration and image segmentation.

\section{Points Clustering}
\label{sec:point:clustering} The neural oscillator points clustering
is based on diffusive connections (\ref{eqn:dif:connection}) and
follows directly the general algorithm in section~\ref{subsec:algo}.
Let us denote $\by_i=(x_i, y_i)$ the coordinates of a point $\bp_i$.
Each point $\bp_i$ is associated to an oscillator $\bx_i$. The
proximity gestalt
principle~\cite{wertheimer23untersuchungen,kanizsa1996gvm,metzger2006ls}
suggests strong coupling between oscillators corresponding to
proximate points. More precisely, the coupling strength between
$\bx_i$ and $\bx_j$ is
\begin{equation} \label{eqn:coupling:strength:clustering}
k_{ij} = \left\{
\begin{array}{cc}
e^{\frac{-|\by_i-\by_j|^2}{\beta^2}} &~~~~\textrm{if}~~j \in \mathcal{N}_i\\
0 &~~~~\textrm{else} \\
\end{array}
\right.,
\end{equation}
where $\mathcal{N}_i$ is a neighborhood of $\bp_i$. For example
$\mathcal{N}_i$ can be defined as the set of $M$ points closest to
$\bp_i$. Then (\ref{eqn:coupling:strength:clustering}) couples an
oscillator $\bx_i$ and with its $M$ nearest neighbors. The local
coupling can propagate to make the coupling in a larger scale.
Higher $M$ value reinforces the coupling. The parameter $\beta$
tunes the size of the clusters one expects to detect. The external
input $I_i$ of the oscillators in (\ref{eqn:oscillator:FN}) are set
as uniformly distributed random variables in the appropriate range.

%\begin{figure}[htbp]
%\begin{center}
%\begin{tabular}{c}
%\includegraphics[width=4cm]{./figures/coupling_point_clutering.jpg}
%\\
%\end{tabular}
%\end{center}
%\caption{Diffusive coupling for point clustering. Each oscillator is
%\textit{directly} coupled with its 3 nearest neighbors. (The arrows
%represent the coupling: $\bx_i \rightarrow \bx_j$ means that the
%latter is coupled with the former.). All oscillators are coupled
%through propagation in this example.} \label{fig:coupling:points}
%\end{figure}

\begin{figure}[htbp]
\begin{center}
\begin{tabular}{ccc}
\hspace{-2ex}
\includegraphics[width=4cm]{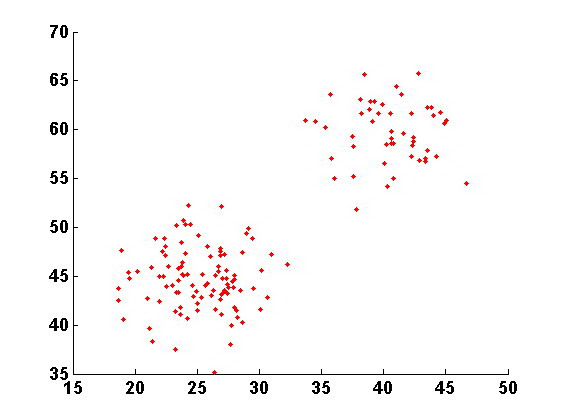}
\hspace{-2ex} &
\includegraphics[width=4cm]{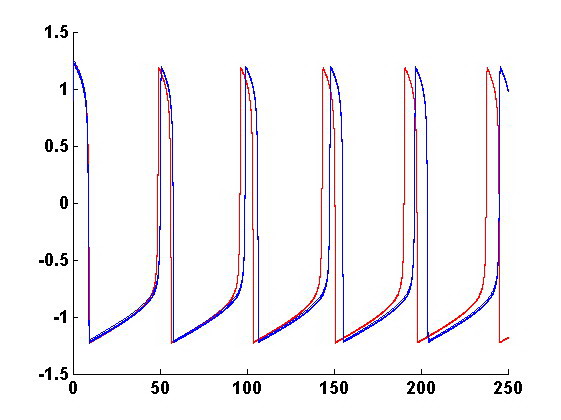}
\hspace{-2ex} &
\includegraphics[width=4cm]{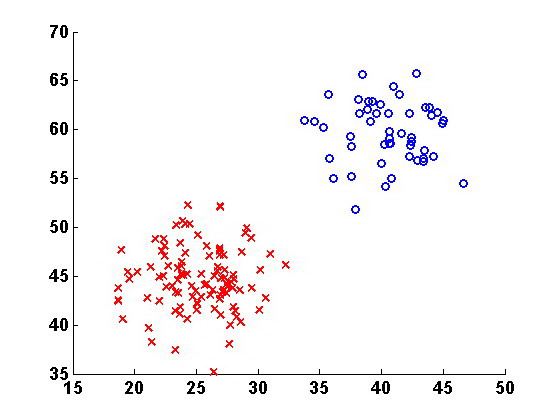}
\\
\textbf{a} & \textbf{b} & \textbf{c} \\
\end{tabular}
\end{center}
\caption{\textbf{a.} Points to cluster. \textbf{b.} The oscillators
converge to two concurrently synchronized groups. \textbf{c.}
Clustering results. The blue circles and the red crosses represent
the two clusters.} \label{fig:clustering:points:2:groups}
\end{figure}

Fig.~\ref{fig:clustering:points:2:groups} illustrates an example in
which the points make clearly two clusters. As shown in
Fig.~\ref{fig:clustering:points:2:groups}-b, the oscillator system
converges to two concurrently synchronized groups, each
corresponding to one cluster, and separated in the time
dimension. The identification of the two groups induces the
clustering of the underlying points, as shown in
Fig.~\ref{fig:clustering:points:2:groups}-c.

\begin{figure}[htbp]
\begin{center}
\begin{tabular}{cc}
\includegraphics[width=4cm]{./figures/1cluster_input.jpg} &
\includegraphics[width=4cm]{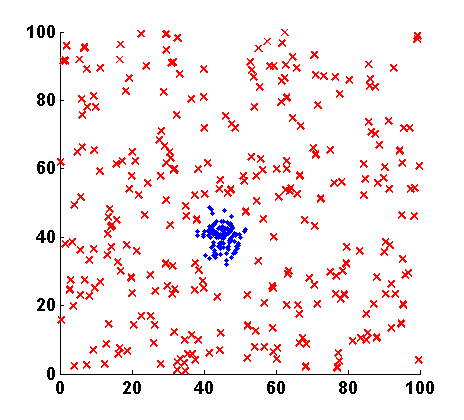} \\
\textbf{a} & \textbf{b} \\
\includegraphics[width=4cm]{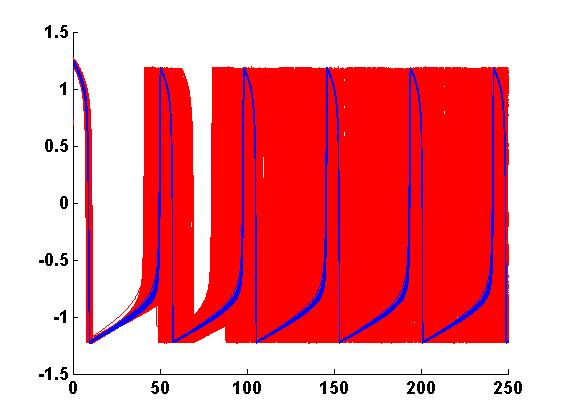} &
\includegraphics[width=4cm]{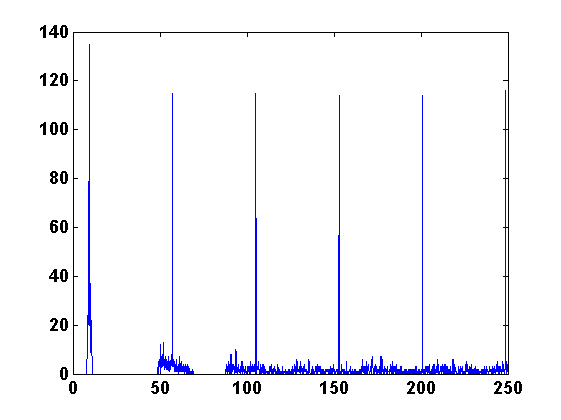} \\
\textbf{c} & \textbf{d}
\\
\end{tabular}
\end{center}
\caption{\textbf{a.} A cloud of points made of 300 points uniformly
randomly distributed in a space of size $100 \times 100$, in
addition to a cluster of 100 Gaussian distributed points with
standard deviation equal to $3 \times 3$. \textbf{b.} Blue dots
represent the cluster detected by the algorithm and red crosses are
the ``outliers''. \textbf{c.} The neural oscillator system converges
to one synchronized group that corresponds to the cluster with all
the ``outliers'' totally desynchronized in the background.
\textbf{d.} The number of oscillators simultaneously spiking.
The peaks in the trace tell the existence of a cluster and their
amplitude (about 115) indicates the number of points in the
cluster.} \label{fig:clustering:points:single:cluster}
\end{figure}

Fig.~\ref{fig:clustering:points:single:cluster} presents a more
challenging setting where one seeks to identify a cluster in a cloud
of points. The cloud is made of 300 points uniformly randomly
distributed in a space of size $100 \times 100$, in addition to a
cluster of 100 Gaussian distributed points with standard deviation
equal to $3 \times 3$. Thanks to the coupling
(\ref{eqn:coupling:strength:clustering}), the neural oscillator
system converges to one synchronized group that corresponds to the
cluster with all the ``outliers'' totally desynchronized in the
background, as shown in
Fig.~\ref{fig:clustering:points:single:cluster}-c. The synchronized
traces are segregated from the background (for example by
thresholding the correlation among the traces) which results in the
identification of the underlying cluster, as shown in
Fig.~\ref{fig:clustering:points:single:cluster}-b.
Fig.~\ref{fig:clustering:points:single:cluster}-d plots along time
the number of the oscillators simultaneously spiking. The peaks in
the trace tell the existence of a cluster (similarly to e.g.
~\cite{hopfield2001mts,wang2006fcn}), and their amplitude (about
115) indicates the number of points in the cluster: the oscillators
which belong to the cluster are synchronized and thus spike together
to make a high peak.

\section{Contour Integration}
\label{sec:contour:integration} 
\begin{figure}[htbp]
\begin{center}
\begin{tabular}{c}
\includegraphics[width=8cm]{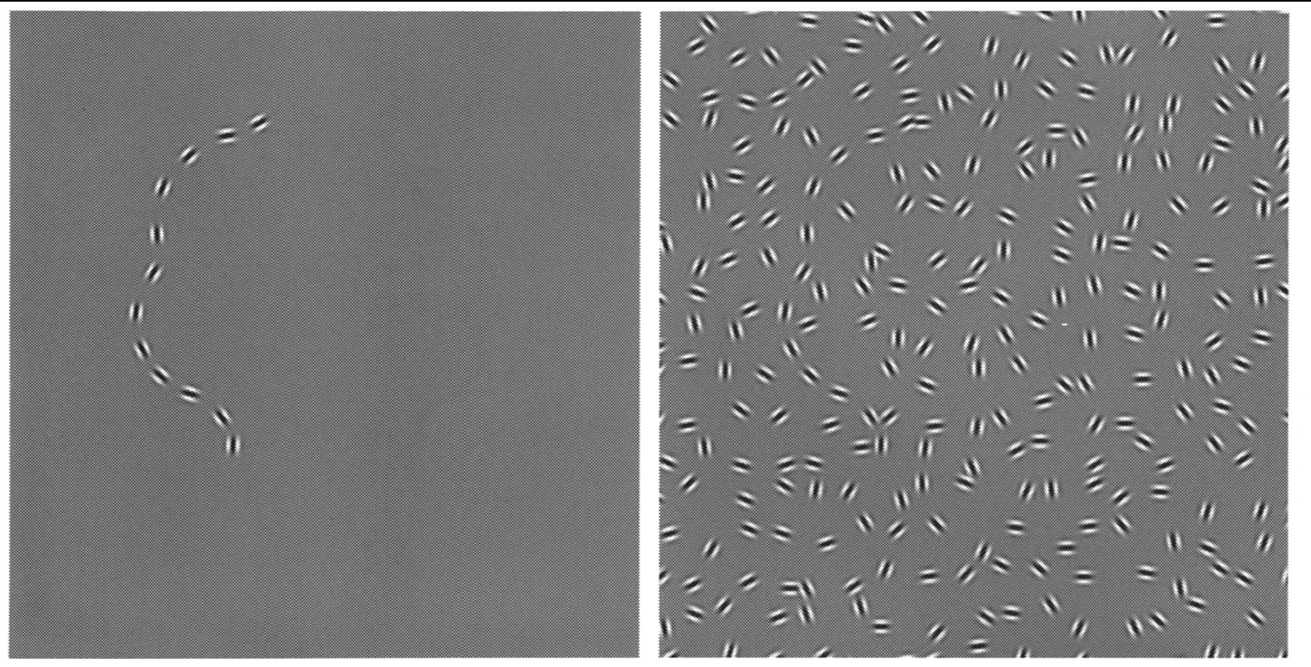} \\
\end{tabular}
\end{center}
\caption{The left-hand panel shows the path of elements (the
stimulus) that the subjects must detect when embedded in an array of
randomly oriented elements (the stimulus plus background shown on
the right). The stimulus consisted of 12 elements aligned along a
path. In this example each successive element differs in orientation
by $30^{\circ}$ and for this difference in orientation the string of
aligned elements is easily detected. This figure is cited
from~\cite{Field93}.} \label{fig:contour:integration:Field}
\end{figure}

\begin{figure}[htbp]
\begin{center}
\begin{tabular}{c}
\includegraphics[width=6cm]{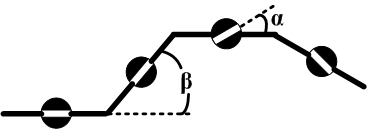} \\
\end{tabular}
\end{center}
\caption{The element-to-element angle $\beta$ is the difference in
angle of orientation of each successive path segment. The
element-to-path angle $\alpha$ is the angle of orientation of the
element with respect to the path.}
\label{fig:contour:integration:angles}
\end{figure}

Field and his
colleagues~\cite{Field93} have shown some interesting experiments,
an example being illustrated in
Fig.~\ref{fig:contour:integration:Field}, to test human capacity of
contour integration, i.e. of identifying a path within a field of
randomly-oriented elements and made some quantitive observations in
accordance with the ``good continuation''
law~\cite{wertheimer23untersuchungen,kanizsa1996gvm,metzger2006ls}:
\begin{itemize}
\item Contour integration can be made when the successive elements in the path,
i.e., the element-to-element angle $\beta$ (see
Fig.~\ref{fig:contour:integration:angles}), differ by $60^\circ$ or
less.
\item There is a constraint between the element-to-element angle
$\beta$ and the element-to-path angle $\alpha$ (see
Fig.~\ref{fig:contour:integration:angles}). The visual system can
integrate large differences in element-to-element orientation only
when those differences lie along a smooth path, i.e., only when the
element-to-path angle $\alpha$ is small enough. For example, with
$\alpha=15^\circ$ and $\beta=0^\circ$ the contour integration is
difficult, even though the observers can easily track a 15 degree
orientation difference when there is no variation ($\alpha=0^\circ$
and $\beta=15^\circ$).
%\item An ``association field'' model is then proposed as illustrated in
%Fig.~\ref{fig:contour:integration:association}.
\end{itemize}

%\begin{figure}[htbp]
%\begin{center}
%\begin{tabular}{c}
%\includegraphics[width=8cm]{./figures/contour_integration_association.jpg} \\
%\end{tabular}
%\end{center}
%\caption{The association field. The diagram at the top of the
%figure (a) represents the rules by which the elements in the path
%are associated and segregated from the background. The curves in (b)
%represent the specific rules of alignment. Grouping occurs only when
%the orientation of elements conforms to first-order curves (i.e.
%curves with no points of inflection) like those shown by the rays
%extending from the center of the elements. The integration process
%thus appears to show strong joint constraints of position and
%orientation. Thus elements with alignment like that shown on the
%bottom left will be associated while elements like that shown on the
%right will not even though the difference in orientation is the same
%in both examples. The figure is cited
%from~\cite{wertheimer23untersuchungen}}
%\label{fig:contour:integration:association}
%\end{figure}

\begin{figure}[htbp]
\begin{center}
\begin{tabular}{ccc}
\includegraphics[width=4cm]{./figures/contour_straight_input1.jpg} &
\includegraphics[width=4cm]{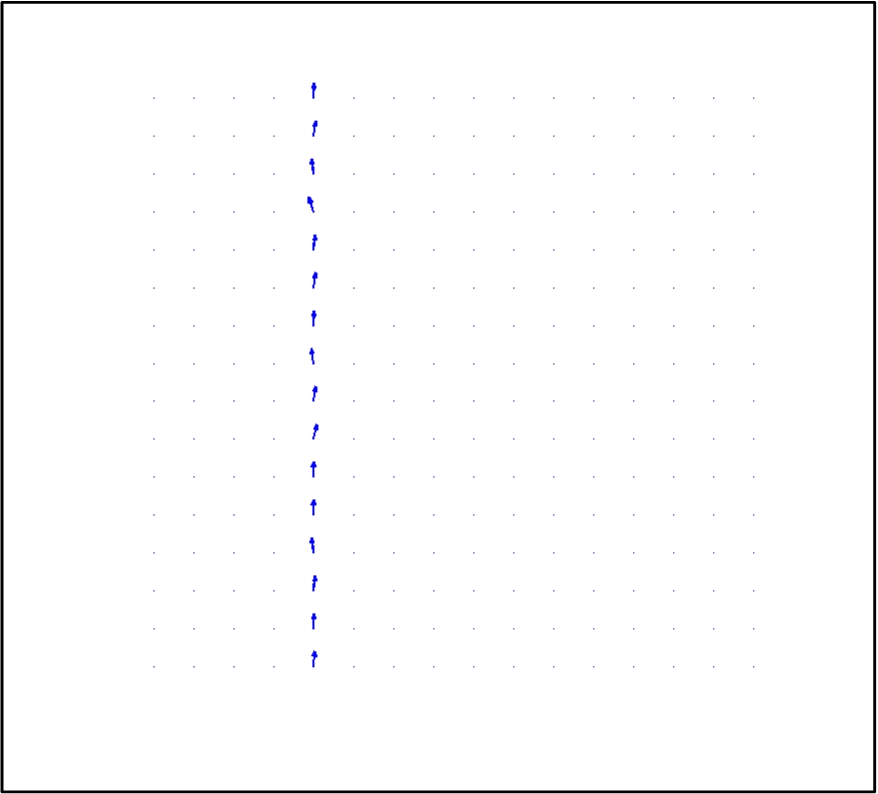} &
\includegraphics[width=4cm]{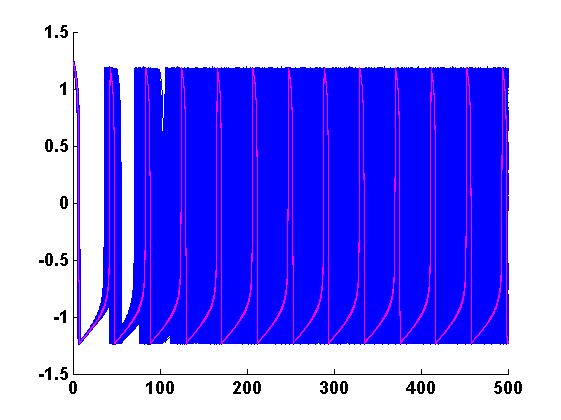}
 \\
 a & b & c \\
\end{tabular}
\end{center}
\caption{Left: A vertical contour is embedded in a uniformly
distributed orientation grid. Middle: The detected contour. Right:
the traces of the neural oscillation.} \label{fig:contour:line}
\end{figure}

Fig.~\ref{fig:contour:line} shows the setting of the contour
integration experiments. An orientation value $\bo_{\bi} \in [0,
2\pi)$ is defined for each point $\bi=(i,j)$ in a grid as
illustrated by the flashes. The proposed algorithm detects the
smooth contours potentially imbedded in the grid.

Following the general visual grouping algorithm described in
section~\ref{subsec:algo}, neural oscillators with diffusive
connections (\ref{eqn:dif:connection}) are used to perform contour
integration. Each orientation in the grid is associated to one
oscillator. The coupling of the oscillators $\bi$ and $\bj$ follows
the Gestalt law of ``good continuation'' and, in particular, the
results of the psychovisual experiments of Field et
al~\cite{Field93}:
\begin{equation} \label{eqn:coupling:strength:contour}
k_{\bi\bj} = \left\{
\begin{array}{cc}
\exp{\left(-\frac{|\bo_{\bi}-\bo_{\bj}|^2}{\delta^2} -
\frac{|\frac{\bo_{\bi}+\bo_{\bj}}{2} - \bo_{\bi\bj}|^2
}{\gamma^2} \right)}  & ~\textrm{if}~|\bi-\bj| \leq w\\
0 &~\textrm{if}~|\bi-\bj|>w \\
\end{array}
\right. .
\end{equation}
where $$\bo_{\bi\bj} = \left\{
\begin{array}{cc}
\arctan(\bi-\bj) & \textrm{if}~|\arctan(\bi-\bj)-\frac{|\bo_{\bi}+\bo_{\bj}|}{2}| < |\arctan(\bi-\bj) + \pi -\frac{|\bo_{\bi}+\bo_{\bj}|}{2}| \\
\arctan(\bi-\bj) + \pi & \textrm{else} \\
\end{array}
\right.
$$ is the undirectional orientation of the path $\bi\bj$ (the closer to the average
element-to-element orientation $\frac{|\bo_{\bi}+\bo_{\bj}|}{2}$
modulo $\pi$). By making constraints on the element-to-element angle
$\beta$ (the first term in (\ref{eqn:coupling:strength:contour}))
and the element-to-path angle $\alpha$ (the second term in
(\ref{eqn:coupling:strength:contour})), the neural oscillator system
makes smooth contour integration. $\delta$ and $\gamma$ tune the
smoothness of the detected contour. As contour integration is known
to be rather local~\cite{Field93}, the coupling
(\ref{eqn:coupling:strength:contour}) is effective within a
neighborhood of size $(2w+1) \times (2w+1)$.

\begin{figure}[htbp]
\begin{center}
\begin{tabular}{ccc}
\includegraphics[width=4cm]{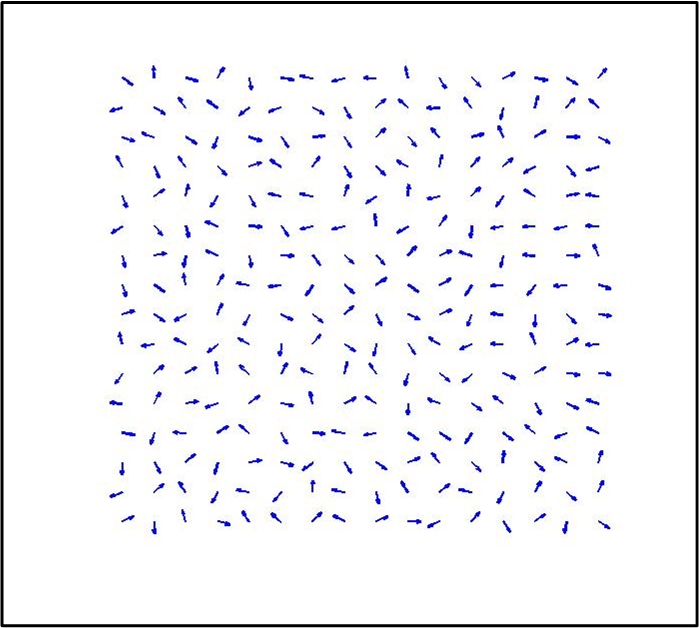} &
\includegraphics[width=4cm]{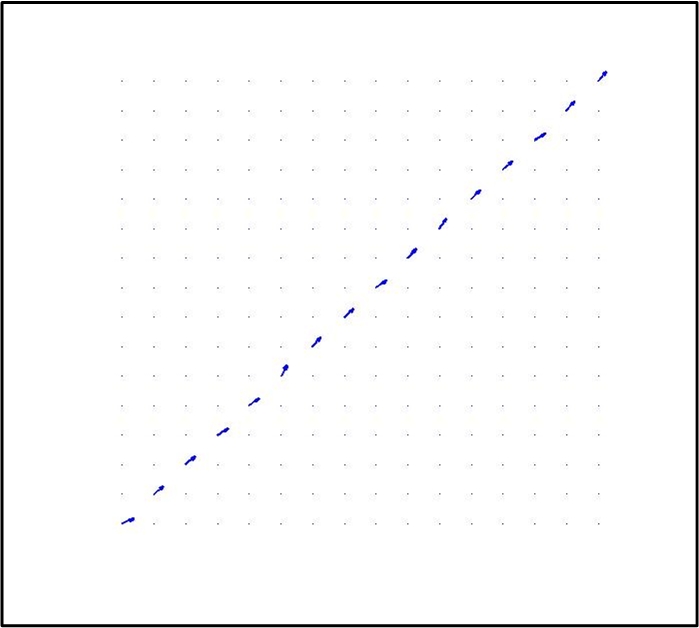} &
\includegraphics[width=4cm]{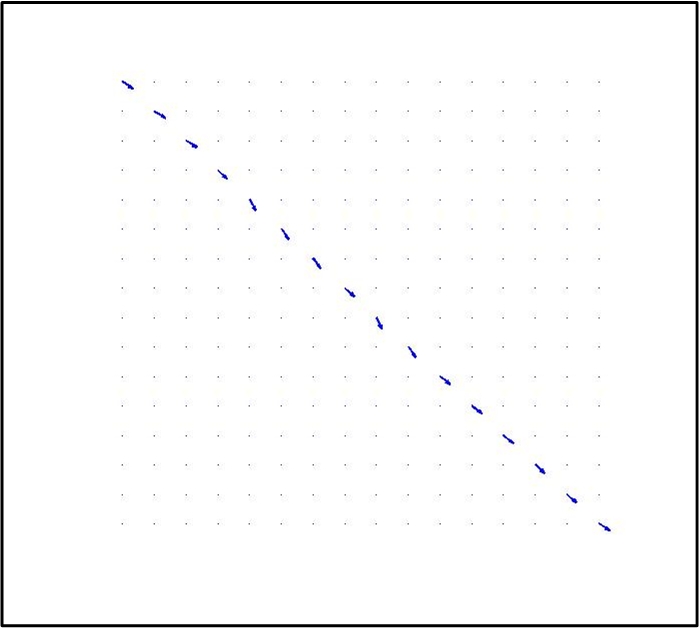}
 \\
 a & b & c \\
\end{tabular}
\end{center}
\caption{Left: Two contours are embedded in a uniformly distributed
orientation grid. Middle and right: the two  identified contours.}
\label{fig:contour:2lines}
\end{figure}

The example illustrated in Fig.~\ref{fig:contour:line}-a presents a
grid in which orientations are uniformly distributed in space,
except for one vertical contour. The orientation of the elements on
the vertical contour undertakes furthermore a Gaussian perturbation
of standard deviation $\sigma=10^\circ$. The neural oscillator
system converges to one synchronized group that corresponds to the
contour with all the other oscillators desynchronized, as
illustrated in Fig.~\ref{fig:contour:line}-c. The synchronized group
is segregated by thresholding the correlation among the traces. This
results in the ``pop-out'' of the contour, as shown in
Fig.~\ref{fig:contour:line}-b. The parameters are configured as
$\delta=20^\circ$, $\gamma=10^\circ$ and $w=1$, which are in line
with the results of the psychovisual experiments of Field et
al~\cite{Field93}. Fig.~\ref{fig:contour:2lines} illustrates a
similar example with two intersected straight contours.

\begin{figure}[htbp]
\begin{center}
\begin{tabular}{cc}
\includegraphics[width=6cm]{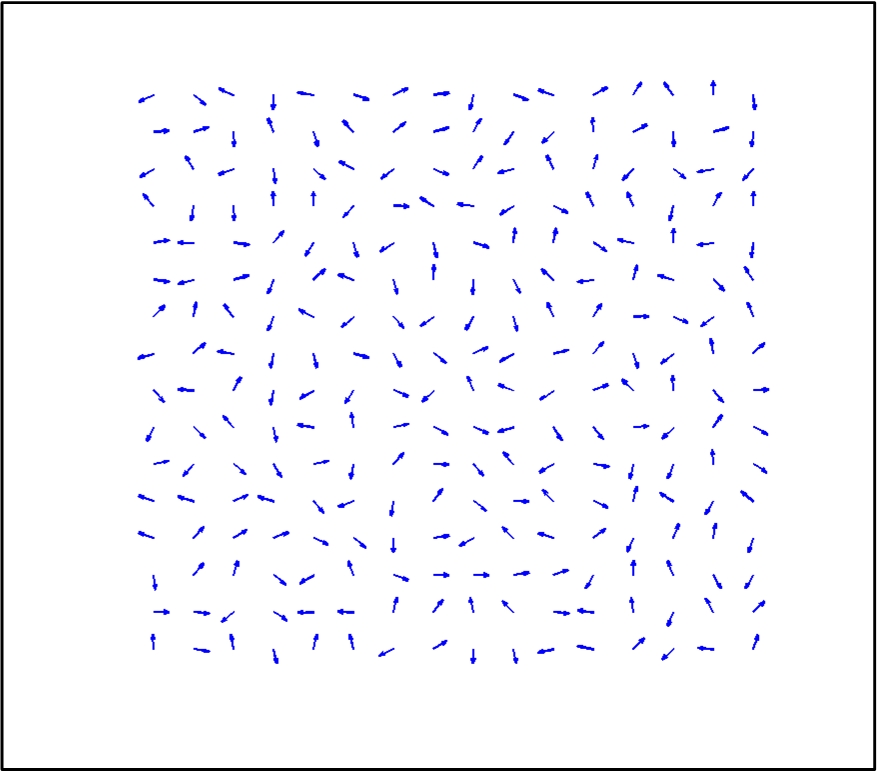} &
\includegraphics[width=6cm]{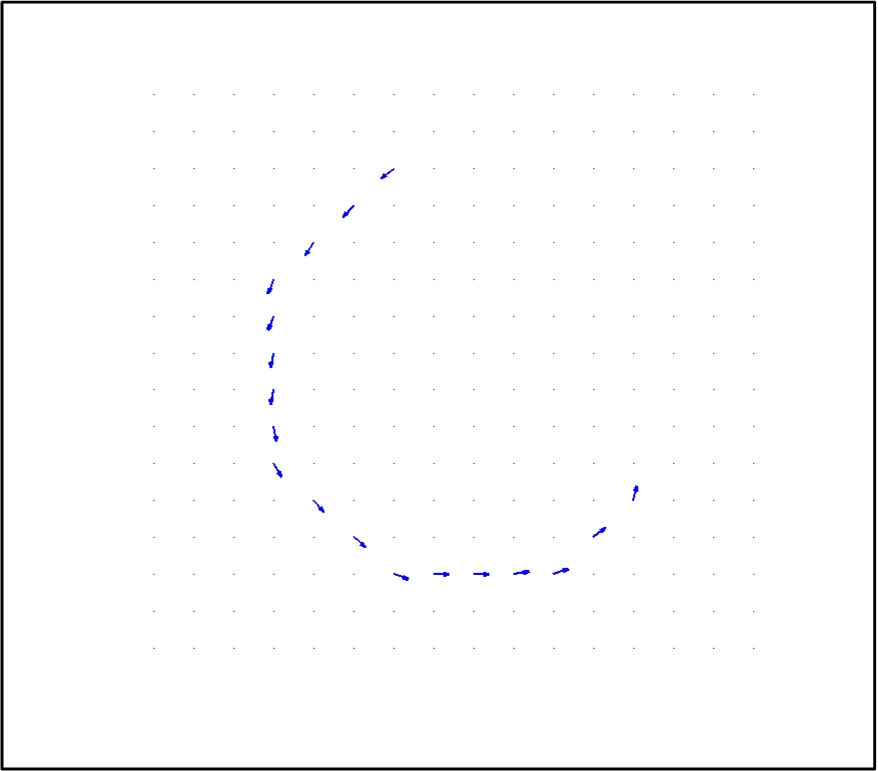}
 \\
 a & b \\
\end{tabular}
\end{center}
\caption{Left: A smooth curve is embedded in a uniformly distributed
orientation grid. Middle: The detected curve. }
\label{fig:contour:curve}
\end{figure}

Fig.~\ref{fig:contour:curve}-a illustrates a smooth curve embedded
in the uniformly randomly distributed orientation background. With
some minor effort, subjects are able to identify the curve due to
its ``good continuation''. Similarly the neural system segregates
the curve from the background with the oscillators lying on the
curve fully synchronized, as illustrated in
Fig.~\ref{fig:contour:curve}-b.

\section{Image Segmentation}
\label{sec:image:segmentation} The proposed image segmentation
scheme is based on concurrent synchronization~\cite{Pham07} and
follows the general visual grouping algorithm described in
section~\ref{subsec:algo}.  In the basic version, the coupling gain
between oscillators are again inspired directly from more standard
techniques, namely non-local grouping as applied e.g.~to in image
denoising~\cite{Buades05, Buades07} in addition to the gestalt laws.
Multi-layer neural networks and feedback mechanisms are then
introduced to reinforce robustness under strong noise perturbation
and to aggregate the grouping. Experiments on both synthetic and
real images are shown.

\subsection{Basic Image Segmentation}

One oscillator is associated to each pixel in the image.  Within a
neighborhood the oscillators are non-locally coupled with a coupling
strength
\begin{equation} \label{eqn:coupling:strength:segmentation}
k_{ij} = \left\{
\begin{array}{cc}
e^{\frac{-|u_{\bi}-u_{\bj}|^2}{\beta^2}} & \textrm{if $|\bi - \bj| < w$}\\
0 & \textrm{otherwise} \\
\end{array}
\right. .
\end{equation}
where $u_{\bi}$ is the pixel gray-level at coordinates $\bi = (i,j)$
and $w$ adjusts the size of the neighborhood. Pixels with similar
grey-levels are coupled more tightly, as suggested by the color
constancy gestalt
law~\cite{wertheimer23untersuchungen,kanizsa1996gvm,metzger2006ls}.
Non-local coupling plays an important role in regularizing the image
segmentation, with a larger $w$ resulting in more regularized
segmentation and higher robustness to noise.

\begin{figure}[htbp]
\begin{center}
\begin{tabular}{ccc}
\includegraphics[width=4cm]{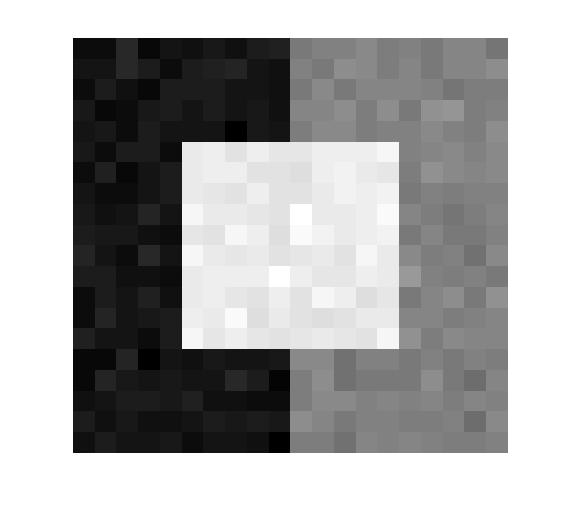} &
\includegraphics[width=4cm]{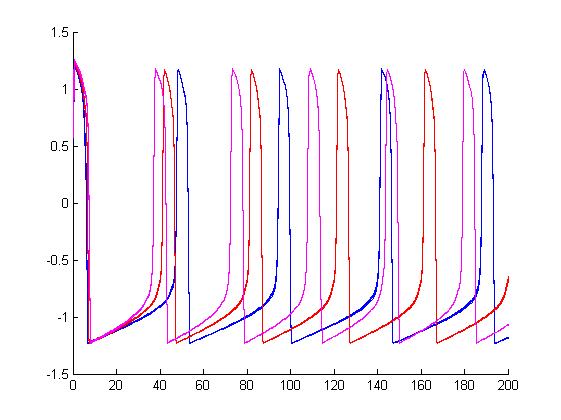} &
\includegraphics[width=4cm]{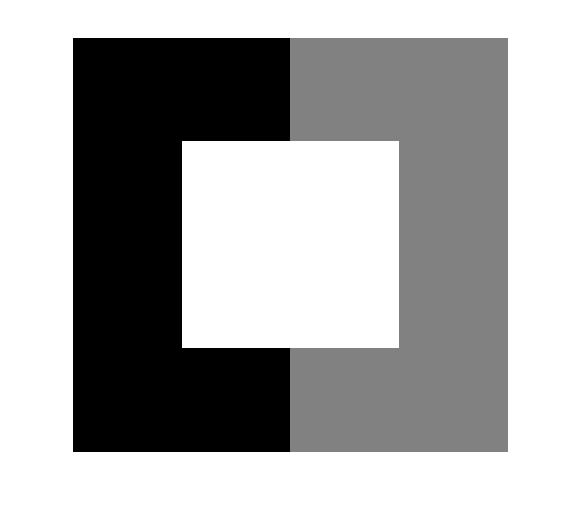} \vspace{-1ex} \\
\textbf{a} & \textbf{b} & \textbf{c} \\
\end{tabular}
\end{center}
\caption{\textbf{a.} A synthetic image (the gray-levels of the
black, gray and white parts are respectively 0, 128, 255)
contaminated by white Gaussian noise of standard deviation
$\sigma=10$. \textbf{b.} The traces of the neural oscillation. The
oscillators converge into three concurrently synchronized groups.
\textbf{c.} Segmentation result.}
\label{fig:segmentation:synthetic:sigma10}
\end{figure}

Fig.~\ref{fig:segmentation:synthetic:sigma10}-a illustrates a
synthetic image (the gray-levels of the black, gray and white parts
are 0, 128, and 255) contaminated by white Gaussian
noise of moderate standard deviation $\sigma=10$. The segmentation
algorithm was configured with $\beta=\sigma$ and $w=5$. The
oscillators converge into three concurrently synchronized groups as
plotted in Fig.~\ref{fig:segmentation:synthetic:sigma10}-b which
results in a perfect segmentation as shown in
Fig.~\ref{fig:segmentation:synthetic:sigma10}-c.

\begin{figure}[htbp]
\begin{center}
\begin{tabular}{cc}
\includegraphics[width=6cm]{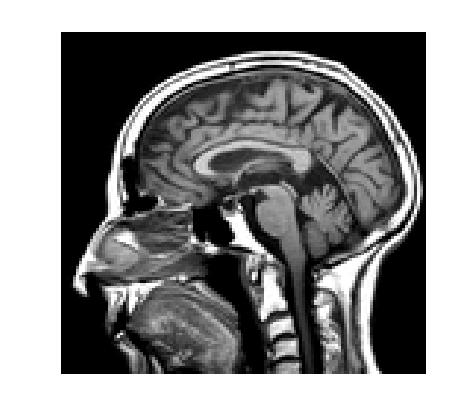} &
\includegraphics[width=6cm]{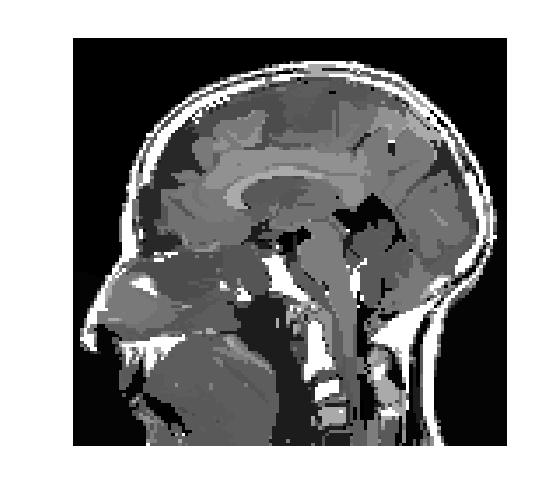} \\
\textbf{a} & \textbf{b} \\
\includegraphics[width=6cm]{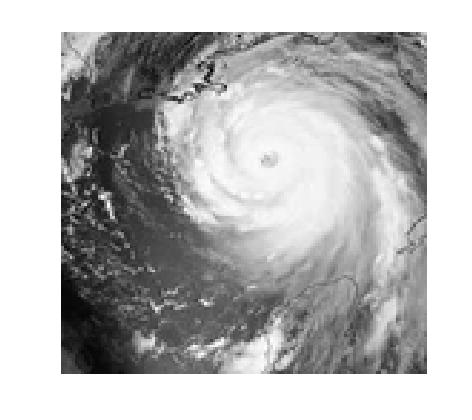} &
\includegraphics[width=6cm]{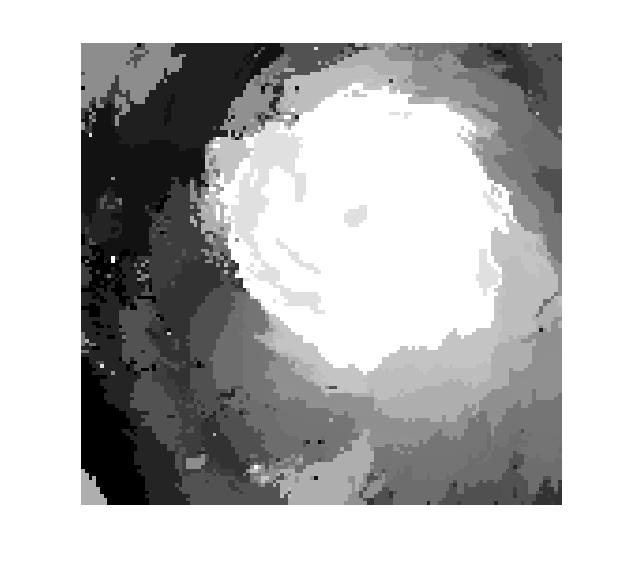} \\
\textbf{c} & \textbf{d} \\
\end{tabular}
\end{center}
\caption{Real image segmentation. \textbf{a} and \textbf{b.} A
sagittal MRI image of size $128 \times 128$ and the segmentation
result in 15 classes. \textbf{c} and \textbf{d.} A radar image of
size $128 \times 128$ and the segmentation result in 20 classes.}
\label{fig:segmentation:katrina}
\end{figure}

Fig.~\ref{fig:segmentation:katrina} show some natural image
segmentation examples. The segmentation results are rather regular
with hardly any ``salt and pepper'' holes, thanks to the
diffusive coupling. A sagittal MRI (Magnetic Resonance Imaging)
image in Fig.~\ref{fig:segmentation:katrina}-a is segmented in 15
classes, the segmentation results shown in
Fig.~\ref{fig:segmentation:katrina}-b. Salient regions such as
cortex, cerebellum and lateral ventricle are segregated with good
accuracy. Fig.~\ref{fig:segmentation:katrina}-c is a radar image in
which boundaries are blurred. In the segmentation results with 20
classes as shown in Fig.~\ref{fig:segmentation:katrina}-d, the
image, including the eye of the hurricane, is accurately
segregated.

\subsection{Noisy Image Segmentation with Feedback}

\begin{figure}[htbp]
\begin{center}
\begin{tabular}{ccc}
\includegraphics[width=4cm]{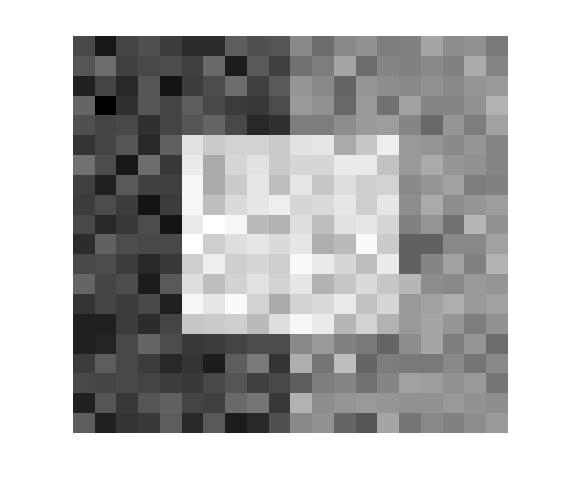} &
\includegraphics[width=4cm]{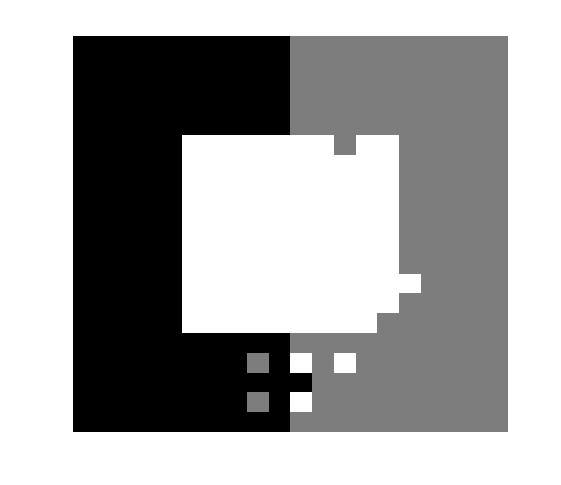} &
\includegraphics[width=4cm]{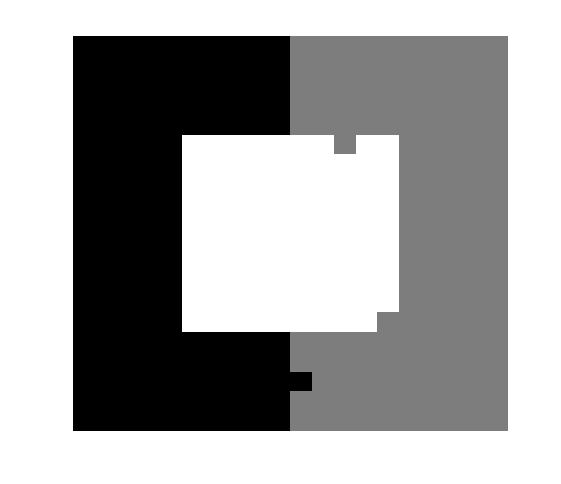} \vspace{-1ex} \\
\textbf{a} & \textbf{b} & \textbf{c} \\
\includegraphics[width=4cm]{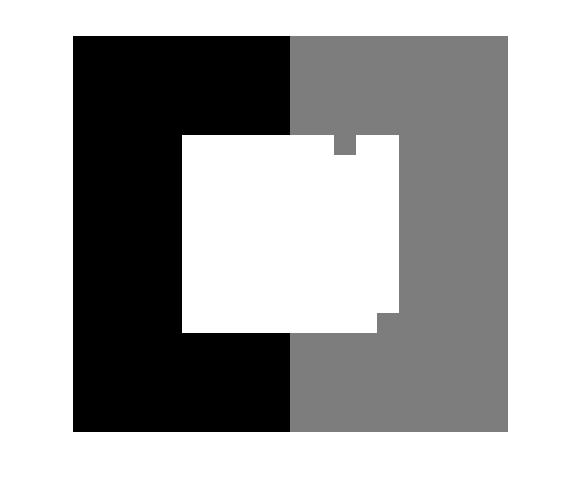} &
\includegraphics[width=4cm]{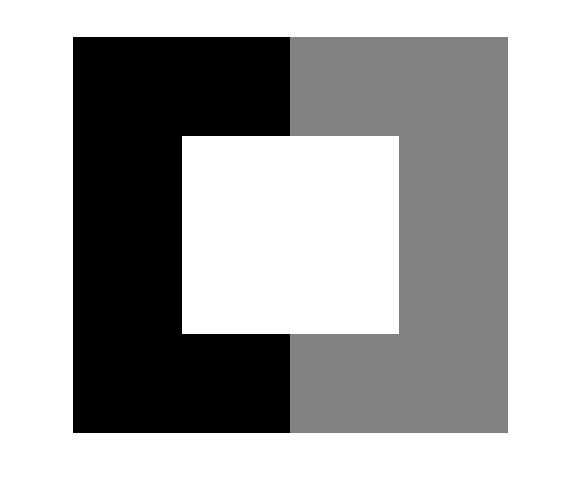} &
\includegraphics[width=4cm]{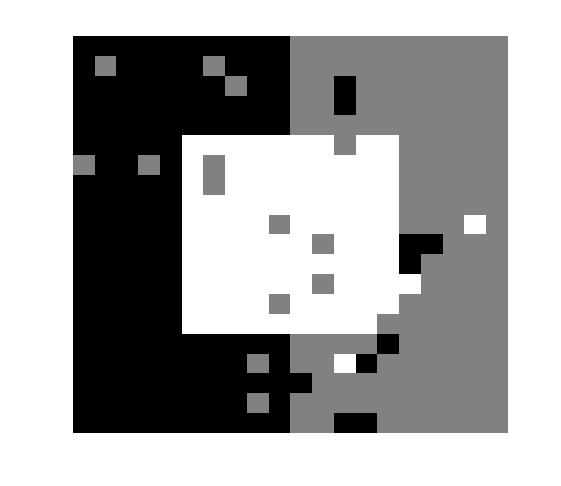} \vspace{-1ex}\\
\textbf{d} & \textbf{e} & \textbf{f} \\
\end{tabular}
\end{center}
\caption{\textbf{a.} A synthetic image (the gray-levels of the
black, gray and white parts are respectively 0, 128, 255)
contaminated by white Gaussian noise of standard deviation
$\sigma=40$. \textbf{b}--\textbf{e.} Segmentation by neural
oscillators with feedback, shown at the 4th, 6th, 8th and 10th
oscillation periods. \textbf{f.} Segmentation by $k$-means.}
\label{fig:segmentation:synthetic}
\end{figure}

Fig.~\ref{fig:segmentation:synthetic}-a shows a synthetic image
contaminated by strong white Gaussian noise of standard deviation
$\sigma=40$. Thanks to the non-local coupling, the neural oscillator
network is robust and segmentation results
(Fig.~\ref{fig:segmentation:synthetic}-b) are more regular than those
of algorithms which do not take into account the image regularity
prior, such as e.g. $k$-means
(Fig.~\ref{fig:segmentation:synthetic}-f). However, due to the heavy
noise perturbation, the segmentation result is not perfect. A feedback
scheme is introduced to overcome this problem. Specifically, in a
loose analogy with the visual cortex hierarchy, a second layer is
introduced to reflect prior knowledge, in this case that proximate
pixels of natural images are likely to belong to the same
region. Feedback from the second layer to the first exploits this
regularity to increase robustness to noise.

\begin{figure}[htbp]
\begin{center}
\begin{tabular}{ccc}
\includegraphics[width=8cm]{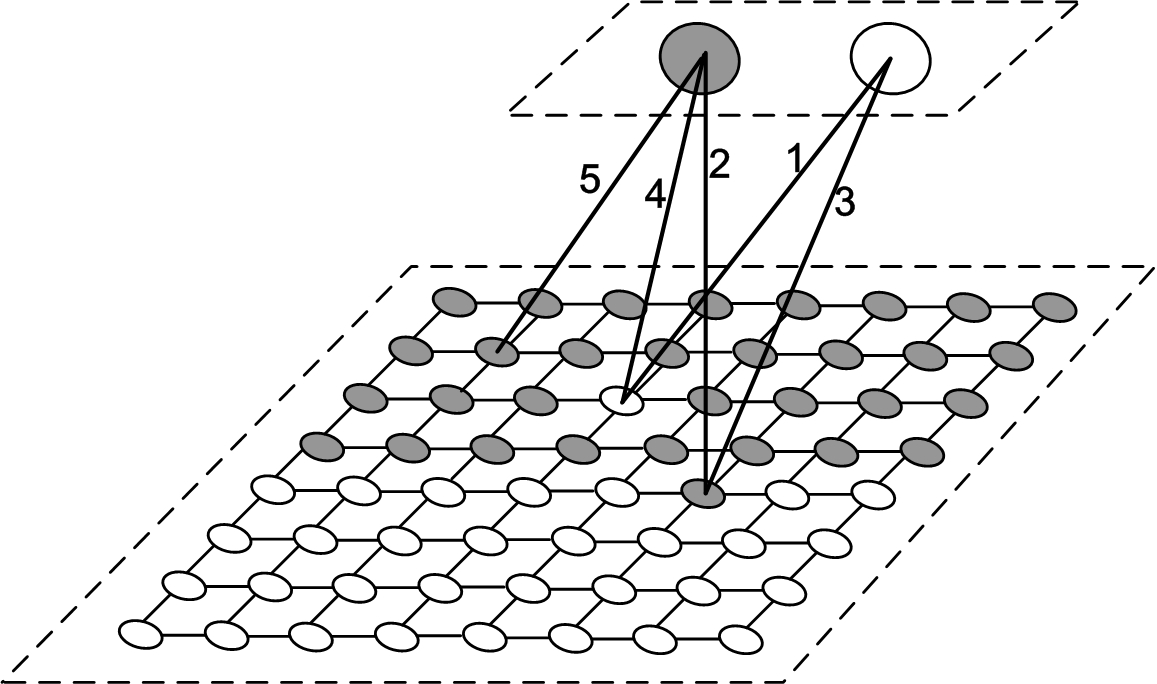} \\
\end{tabular}
\end{center}
\caption{Neural network with feedback for image segmentation. In the
first layer one oscillator is associated to each pixel. An
oscillator is coupled within all the others in a neighborhood of
size $(2w+1)\times(2w+1)$ (for clarity only the coupling
with the 4 nearest neighbors are shown in the figure). The image is
segmented into two regions marked by white and gray. The second
layer contains two oscillators whose input are respectively the
average gray-level of the two image regions. The coupling strength
between an oscillator indexed by the coordinates $(i,j)$ in the
first layer and an oscillator indexed by the regions $m=1,2$ in the
second layer is proportional to number of the pixels in the
neighborhood $\mathcal{N}(i,j)=\{(i,j), (i\pm1,j), (i,j\pm1)\}$ that
belong to the region $m$ according to the previous segmentation.
Only five such couplings and their strengths are shown for clarity
.} \label{fig:image:seg:feedback}
\end{figure}

The feedback error correction mechanism is implemented using the
generalized diffusive connections introduced in
Section~\ref{subsubsec:generalized}. On top of the first layer
previously described, a second layer of oscillators is added and
coupled with the first layer, as shown in
Fig.~\ref{fig:image:seg:feedback}. The second layer contains $M$
oscillators, where $M$ is the number of regions obtained in the first
layer, the input of each oscillator being the average gray-level of an
image region. Each oscillator in the first layer, indexed by $(i,j)$,
is coupled with all the $M$ oscillators in the second layer, with the
coupling strengths depending on the segmentation obtained in a
neighborhood of $(i,j)$.  More precisely, the coupling matrices $\bA$
and $\bB$ in the generalized diffusive connection
(\ref{eqn:generalized:l1}) are designed so that the coupling strength
from an oscillator $\bx_2(m)$, $m=1, \ldots, M$, in the second layer
to an oscillator $\bx_1(i,j)$ in the first layer is proportional to
number of pixels in the neighborhood $\mathcal{N}(i,j)$ which belong
to the region $m$ according to the current segmentation $-$ as
illustrated in Fig.~\ref{fig:image:seg:feedback}. This inter-layer
connection reinforces the coupling between the first-layer oscillators
and the second-layer oscillators which correspond to the locally
dominant regions, and thus regularizes the segmentation. After each
step of $\bA$ and $\bB$ adjustment, the oscillator network converges
to a new concurrently synchronized status and the segmentation is
updated. Figs.~\ref{fig:segmentation:synthetic}-b
to~\ref{fig:segmentation:synthetic}-e show the segmentation results at
the beginning of the 4th, 6th, 8th and 10th oscillation periods. The
segmentation errors are corrected as the feedback is going along. The
segmentation is stable at Fig.~\ref{fig:segmentation:synthetic}-e
after the 10th period.

\begin{figure}[htbp]
\begin{center}
\begin{tabular}{ccc}
\hspace{-9ex}
\includegraphics[width=6cm]{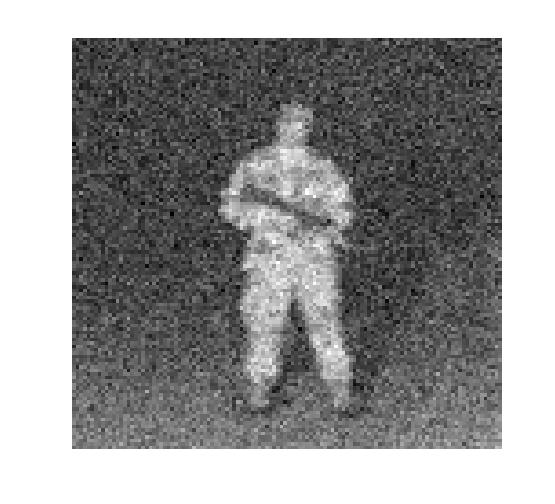} &
\hspace{-7ex}
\includegraphics[width=6cm]{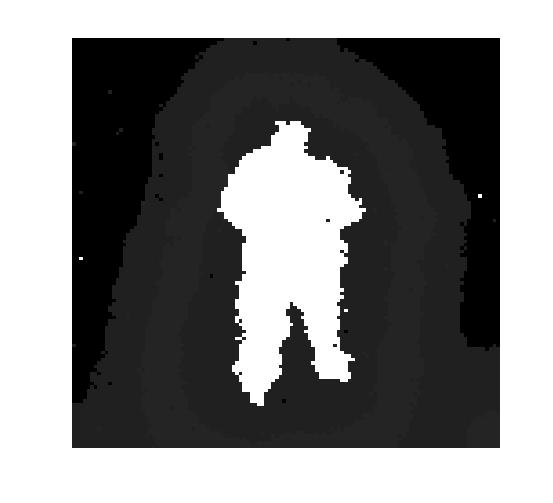} &
\hspace{-7ex}
\includegraphics[width=6cm]{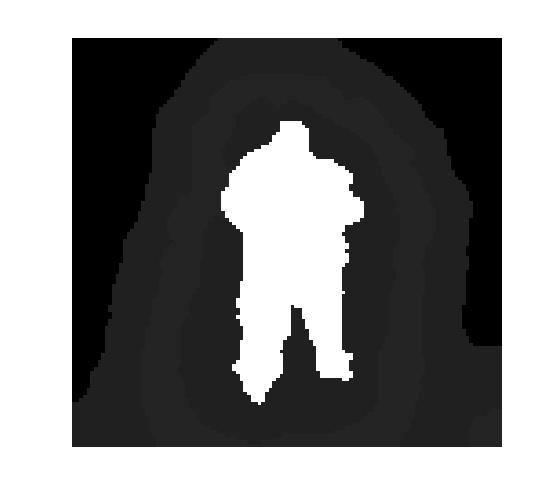} \vspace{-3ex}\\
\hspace{-9ex}\textbf{a} & \hspace{-7ex}\textbf{b} & \hspace{-7ex}\textbf{c} \\
\end{tabular}
\vspace{-3ex}
\end{center}
\caption{Infrared image segmentation. \textbf{a.} Infrared image.
\textbf{b.} Segmentation without feedback. \textbf{c.} Segmentation
with feedback.} \label{fig:seg:infra}
\end{figure}

Fig.~\ref{fig:seg:infra}-a illustrates an infrared night vision
image heavily noised. The segmentation result of the basic algorithm
without feedback shown in Fig.~\ref{fig:seg:infra}-b contains a few
punctual errors and, more importantly, the contour of the segmented
objected zigzags due to the strong noise perturbation. As
illustrated in Fig.~\ref{fig:seg:infra}-c, the feedback procedure
corrected the punctual errors and regularized the contour.

\subsection{Multi-layer Image Segmentation}

The visual cortex is hierarchical, with cells at the lower levels
having smaller reception fields than the ones at the higher levels
and information aggregating from bottom to
top~\cite{rao1999pcv,hawkins2004ihh}. This structure can be imitated
by multi-layer oscillator networks.

\begin{figure}[htbp]
\begin{center}
\begin{tabular}{c}
\includegraphics[width=8cm]{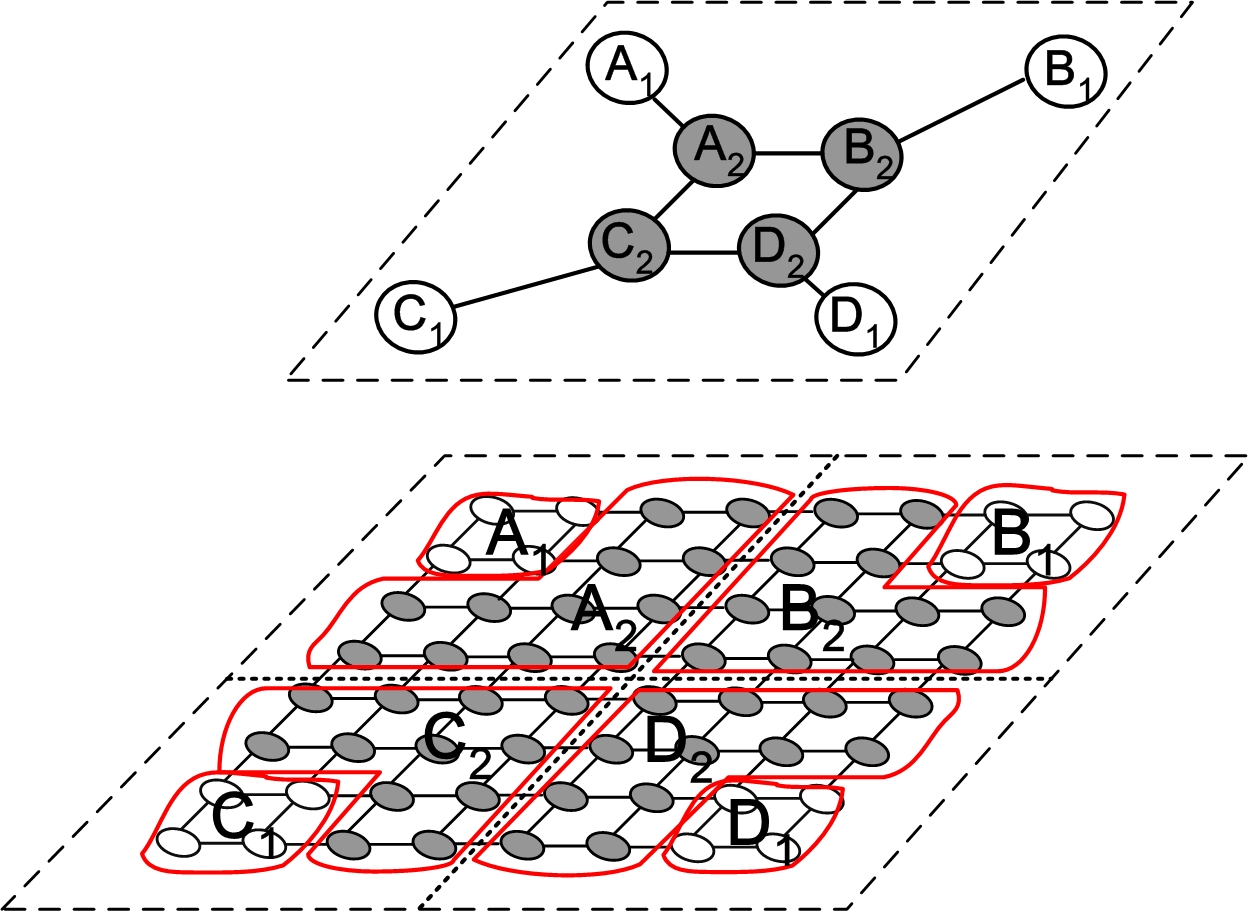}
\end{tabular}
\end{center}
\caption{Two-layer image segmentation.} \label{fig:seg:multilayer}
\end{figure}

Fig.~\ref{fig:seg:multilayer} illustrates a two-layer image
segmentation scheme. The image on the first layer is decomposed into
four disjoint parts. Each part is treated as an independent image
where a basic image segmentation is performed (in this symbolic
example, each part is segmented into two regions.) The second layer
aggregates the segmentation results obtained on the first layer,
using a single oscillator to represent each resulting region, with
the average gray-level in the region as the oscillator input. The
coupling connections on the second layer follow the topology of the
regions on the first layer: second-layer oscillators whose
underlying regions in the first layer are in vicinity are coupled.
The segmentation on the second layer merges some regions obtained in
the first layer and provides the final segmentation result. (In this
example, regions $A_2$, $B_2$, $C_2$ and $D_2$ are merged on the
second layer.) Multi-layer segmentation can follow the same
principle. From a computational point of view, the multi-layer
scheme saves memory and accelerates the program.

\begin{figure}[htbp]
\begin{center}
\begin{tabular}{ccc}
\hspace{-11ex} \includegraphics[width=6cm]{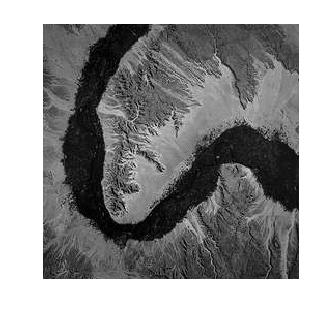}
 & \hspace{-6ex}
\includegraphics[width=6cm]{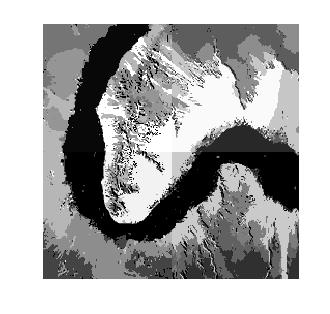} & \hspace{-6ex}
\includegraphics[width=6cm]{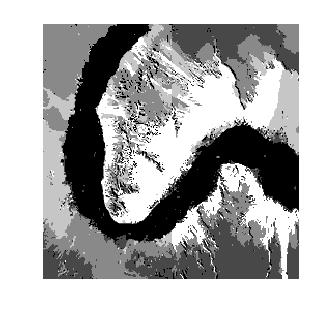} \\
\hspace{-11ex}\textbf{a}  & \hspace{-6ex} \textbf{b} & \hspace{-6ex}\textbf{c} \\
\end{tabular}
\end{center}
\caption{Multi-layer image segmentation. \textbf{a.} Aerial image
($256\times256$). \textbf{b.} Segmentation result of the first
layer. Each of the four $128 \times 128$ parts is segmented in 6
classes. \textbf{c.} Segmentation result in 6 classes with the second
layer.}\label{fig:seg:niler}
\end{figure}

Fig.~\ref{fig:seg:niler} illustrate an example of the multi-layer
image segmentation. The image of size $256 \times 256$ shown in
Fig.~\ref{fig:seg:niler}-a is decomposed into four $128 \times 128$
parts. Fig.~\ref{fig:seg:niler}-b and c present respectively the
segmentation results of the first and second layers, both with 6
classes. From the first layer to the second, some regions that
belong to different parts are merged, which eliminates some boundary
artifacts in the first layer. The merging of a number of regions
belonging to the same part, on the other hand, contributes to the
correction of some over-segmentation. The segmentation result in
Fig.~\ref{fig:seg:niler}-c is satisfactory. The river, although
rather noisy in the original image, is segmented from the land with
accurate boundary. The two aspects of the land are accurately
segregated as well.

\section{Concluding Remarks}
\label{sec:conclusion}

Inspired by neural synchronization mechanisms for perceptual grouping,
simple networks of neural oscillators coupled with diffusive
connections have been proposed to solve visual grouping
problems. Stable multi-layer algorithms and feedback mechanisms have
also been studied.  The same algorithm has been shown to achieve
promising results on several classical visual grouping problems,
including point clustering, contour integration and image
segmentation.

\appendix
  \begin{center}
    \textbf{Appendix: Convergence and Stability}
  \end{center}

One can verify (\cite{Pham07}, to which the reader is referred for more
details on the analysis tools) that a sufficient condition for global
exponential concurrent synchronization of an oscillator network is
\begin{equation}
\lambda_{\min}(\bV \bL \bV^T) > \sup_{\ba, t} \lambda_{\max}\left(
\frac{\partial \bf}{\partial \bx} (\ba, t)\right),
\label{eqn:synch:diffusive}
\end{equation}
where $\lambda_{\min} (\bA)$ and $\lambda_{\max} (\bA)$ are
respectively the smallest and largest eigenvalues of the symmetric
matrix $\bA_s=(\bA + \bA^T)/2$, $\bL$ is the Laplacian matrix of the
network ($\bL_{ii} = \sum_{j \neq i} k_{ij}$, $\bL_{ij} = - k_{ij}$
for $j \neq i$) and $\bV$ is a projection matrix on
$\mathcal{M}^\bot$. Here $\mathcal{M}^\bot$ is the subspace orthogonal
to the subspace $\mathcal{M}$ in which \textit{all} the oscillators
are in synchrony $-$ or, more generally in the case of a hierarchy,
where all oscillators at each level of the hierarchy are in synchrony
(i.e., $\bx_1 = \dots = \bx_N$ at each level). Note that $\mathcal{M}$
itself need not be invariant (i.e. all oscillators synchronized at
each level of the hierarchy need not be a particular solution of the
system), but only needs to be a \textit{subspace} of the actual
invariant synchronization subspace (\cite{lgerard08},~\cite{Pham07}
section 3.3.i.), which may consist of synchronized subgroups according
to the input image. Indeed, the space where all $\bx_i$ are equal (or,
in the case of a hierarchy, where at each level all $\bx_i$ are
equal), while in general not invariant, is always a subspace of the
actual invariant subspace corresponding to synchronized subgroups.

These results can be applied e.g. to individual oscillator dynamics of
the type (\ref{eqn:oscillator:FN}). Let ${\mathbf J_i}$ denote the
Jacobian matrix of the individual oscillator dynamics
(\ref{eqn:oscillator:FN}), ${\mathbf J}_i = \partial [\dot{v}_i \
\dot{w}_i ]^T / \partial [v_i, w_i]^T$. Using a diagonal metric
transformation ${\mathbf \Theta} = {\rm diag}(\sqrt{c \alpha
\beta},1)$, one easily shows, similarly to~\cite{wang05a, wang05b},
that the transformed Jacobian matrix $ {\mathbf \Theta} {\mathbf J}_i
{\mathbf \Theta}^{-1} - {\rm diag}(k, 0) $ is negative definite for $\
k \ > \ 3 + \frac{\alpha \beta}{4} \ $. More general forms of
oscillators can also be used. For instance, other second-order models
can be created based on a smooth function $f(.)$ and an arbitrary
sigmoid-like function $\sigma(.)  \ge 0$ such that $\ 0 \le \sigma'(.)
\le 1 \ $, in the form
\begin{eqnarray}
\label{eqn:oscillator:FN} \dot{v}_i & = & f(v_i) - w_i + I_i \\
\dot{w}_i & = & c [\alpha \sigma(\beta v_i) - w_i]
\end{eqnarray}
with the transformed Jacobian matrix negative definite for
$\ k \ > \ f'(v_i) + \frac{\alpha \beta}{4} \ $.

From a stability analysis point of view, the coupling matrix
${\mathbf{K}}$ of the Gaussian-tuned coupling, composed of coupling
coefficients $k_{ij}$ in Eq.(\ref{eqn:coupling:strength}), presents
desirable properties. It is symmetric, and the classical theorem of
Schoenberg (see~\cite{Micchelli86}) shows that it is positive
definite. Also, although it may actually be state-dependent, the
coupling matrix ${\mathbf{K}}$ can always be treated simply as a
time-varying external variable for stability analysis and Jacobian
computation purposes, as detailed in (\cite{wang05a}, section 4.4,
\cite{Pham07}, section 2.3).

Last, note that more generally the individual dynamics need not all be
oscillators.  In particular, memory-like or voting-like subdynamics
could be introduced in a feedback hierarchy, by having the
corresponding $\mathbf{f}_i (\bx_i, t)$ be the gradient of a scalar
function of $\bx_i$ with multiple local minima.

\bibliographystyle{plain}
\bibliography{these_biblio_final}

\end{document}